\documentclass[sigconf]{acmart}
\usepackage{booktabs} % For formal tables
\usepackage[utf8]{inputenc}
\usepackage{epsfig}
\usepackage{graphicx}
\usepackage{xcolor}
\usepackage{subcaption}
\usepackage{caption}
\usepackage{amsmath}
\usepackage{hyperref}
\usepackage{multirow}
\usepackage{tabularx}
\usepackage[autostyle]{csquotes}
\usepackage{url}
\usepackage{algorithm}
\usepackage{algpseudocode}

\AtBeginDocument{%
  \providecommand\BibTeX{{%
    \normalfont B\kern-0.5em{\scshape i\kern-0.25em b}\kern-0.8em\TeX}}}

\copyrightyear{2022}
\acmYear{2022}
\setcopyright{acmcopyright}
\acmConference[WWW '22 Companion] {Companion Proceedings of the Web Conference 2022}{April 25--29, 2022}{Virtual Event, Lyon, France.}
\acmBooktitle{Companion Proceedings of the Web Conference 2022 (WWW '22 Companion), April 25--29, 2022, Virtual Event, Lyon, France}
\acmPrice{15.00}
\acmISBN{978-1-4503-9130-6/22/04}
\acmDOI{10.1145/XXXXXX.XXXXXX}
\begin{document}
\sloppy\hyphenpenalty=3500
\title{Deep Partial Multiplex Network Embedding}
\author[Q Wang]{Qifan Wang}
\authornote{Corresponding Authors.}
\affiliation{%
  \institution{Facebook AI}
  \city{Menlo Park}
  \state{CA}
  \country{USA}}
\email{wqfcr@fb.com}

\author[Y Fang]{Yi Fang}
\affiliation{%
  \institution{Santa Clara University}
  \city{Santa Clara}
  \state{CA}
  \country{USA}}
\email{yfang@scu.edu}

\author[A Ravula]{Anirudh	Ravula}
\affiliation{%
  \institution{Google Research}
  \city{Mountain View}
  \state{CA}
  \country{USA}}
\email{ravulaanirudh25@gmail.com}

\author[R He]{Ruining He}
\affiliation{%
  \institution{Google Research}
  \city{Mountain View}
  \state{CA}
  \country{USA}}
\email{ruininghe@google.com}

\author[B Shen]{Bin Shen}
\affiliation{%
  \institution{Google Research}
  \city{Mountain View}
  \state{CA}
  \country{USA}}
\email{stanshenbin@gmail.com}

\author[J Wang]{Jingang Wang}
\affiliation{%
  \institution{Meituan NLP Center}
  \city{Beijing}
  \country{China}}
\email{wangjingang02@meituan.com}

\author[X Quan]{Xiaojun	Quan}
\affiliation{%
  \institution{Sun Yat-sen University}
  \city{Guangzhou}
  \country{China}}
\email{quanxj3@mail.sysu.edu.cn}

\author[D Liu]{Dongfang Liu}
\authornotemark[1]
\affiliation{%
  \institution{Rochester Institute of Technology}
  \city{Rochester}
  \state{NY}
  \country{USA}}
\email{dongfang.liu@rit.edu}
\begin{abstract}
Network embedding is an effective technique to learn the low-dimensional representations of nodes in networks.
Real-world networks are usually with multiplex or having multi-view representations from different relations.
Recently, there has been increasing interest in network embedding on multiplex data.
However, most existing multiplex approaches assume that the data is complete in all views. But in real applications, it is often the case that each view suffers from the missing of some data and therefore results in partial multiplex data.

In this paper, we present a novel Deep Partial Multiplex Network Embedding approach to deal with incomplete data. In particular, the network embeddings are learned by simultaneously minimizing the deep reconstruction loss with the autoencoder neural network, enforcing the data consistency across views via common latent subspace learning, and preserving the data topological structure within the same network through graph Laplacian. We further prove the orthogonal invariant property of the learned embeddings and connect our approach with the binary embedding techniques.
Experiments on four multiplex benchmarks demonstrate the superior performance of the proposed approach over several state-of-the-art methods on node classification, link prediction and clustering tasks.
\end{abstract}



% \ccsdesc[500]{Information systems~Network data models}
% \ccsdesc[500]{Computing methodologies~Semantic networks}

% \ccsdesc[500]{Information systems~Network data models}


\maketitle

\section{Introduction}
Network embedding is designed for learning low-dimensional and typically non-linear representations of nodes in the network, which is able to preserve network information. Network embedding has been shown to be useful in many downstream tasks, such as node classification~\cite{WangTAL16}, node clustering~\cite{GaoH18b}, link prediction~\cite{Li0ZZC19} and community detection~\cite{WeiCSLY16}. A variety of network embedding techniques have been proposed in the literature~\cite{DeepWalk,DRNE,wang2017instance,DuLWSW018,WangLDWZ19,ZhangSDYJ19,MengLBZ19,0004ZMK20,BhowmickMDGM20,SunY0CCSH21,Pan0T0021,JiangKS21,LiuHYD21,MAVE,WebFormer}. However, most of these methods focus on single networks, where nodes in the networks are only associated with one type of features or relations.

In many real-world applications, data usually have multiplex representations~\cite{MNE}, where nodes are associated with multiple features from different sources. Multiple types of edges/relations are then generated from these disparate features. For example, in document corpus, a document has hyperlink feature that connects to other related documents. It can also have semantic representation such as attribute or tag feature. Documents are linked together in the attribute view if they share at least one attribute. In Flickr~\cite{FlickrData}, users can be represented with their friendship to others, public comments, photos, reviews, tags, etc. Similarly, users are linked in the photo network or tag network if they share same photos or tags. Previous research on multiplex representation learning~\cite{ICML15,LiuZLTZY20} has demonstrated improved performance by leveraging complementary information from different views. Therefore, there is a growing interest in multiplex network embedding that effectively integrates information from disparate views~\cite{ZhangLSC18,Kawamae19,ZhaoWSLY20}.

Although existing multiplex network embedding methods generate promising results in dealing with multiplex data, most of them assume that all nodes in the network have full information in all views. However, in real-world tasks, it is often the case that a view suffers from some missing information, which results in partial data~\cite{LiJZ14,WangSS15,abs-2003-13088}. For instance, in document corpus, many documents may not contain any hyperlink or tag information. In Flickr, a user might have few or even no friend connections, reviews or tags, resulting in an isolated node in the corresponding relationship network. Moreover, it is also common that users don't share some of their information, such as photos and comments, for privacy consideration. Therefore, it is a practical and important research problem to design effective network embedding methods on partial multiplex data.

There are several ways to apply existing multiplex network embedding methods to partial data. One can either remove the data that suffer from missing information, or preprocess the partial data by first filling in the missing data. The first strategy is clearly not suitable since the purpose is to map all nodes to their corresponding embedding vectors, and our experiments show that the second strategy does not achieve good performance either.
In this paper, we propose a novel Deep Partial Multiplex Network Embedding (DPMNE) approach to deal with such incomplete data. More specifically, a deep autoencoder network is introduced to learn the deep representations of node features. A unified learning framework is developed to learn the network embedding, which simultaneously minimizes the reconstruction error from the deep autoencoder, enforces the data consistency across different views via common latent subspace learning, and preserves data proximity within the same network through graph Laplacian. A coordinate descent algorithm is applied as the optimization procedure. We then further connect our approach to binary embedding methods~\cite{WangZSSS18} based on the orthogonal invariant property of our formulation. Experiments on four multiplex datasets demonstrate the advantages of the proposed approach over
several state-of-the-art single and multiplex network embedding methods.
% We summarize the main contributions of this work as follows:
% \begin{itemize}
% \item We propose a unified network embedding approach to deal with partial multiplex data, which generates effective embedding representations. As far as we know, it is the first attempt to learn network embedding on partial multiplex data.
% \item We propose a coordinate descent method for the joint optimization problem. We prove the orthogonal invariant property of the optimal solution and connect our work to binary embedding methods.
% \item Our extensive experiments demonstrate DPMNE is an effective network embedding method, especially when only partial multiplex data are available.
% \end{itemize}

%rest of the paper
% The rest of the paper is organized as follows. Section 2 reviews the related work.
% Section 3 formally defines our problem and presents the proposed approach. Experimental setting and result are discussed in Section 4. The last section provides conclusions and points out possible future directions.

\section{Related Work}
% Network embedding focuses on generating the low-dimensional vector representation of nodes for real networks or graphs to facilitate further analysis of networks. Traditional approaches to network embedding can be divided into two categories: single network methods and multiplex ones. We review methods in both categories in the following subsections. Moreover, we provide discussions over the existing methods on partial data learning.

\subsection{Single Network Embedding}
Single network embedding methods~\cite{GongLSW20,aaai2021_a,aaai2021_b,feng2021should,chen2021catgcn,wu2022graph} learn an information preserving embedding of a single-view network for node classification, node clustering and many other related tasks. A spectral based method~\cite{BelkinN01} has been proposed, which uses the top-k eigenvectors to represent the network nodes. DeepWalk~\cite{DeepWalk} introduces the idea of Skip-gram to learn node representations from random-walk sequences. LINE~\cite{LINE} tries to use the embedding to approximate the first-order and second-order proximities of the network.
On top of DeepWalk, Node2Vec~\cite{node2vec} adds two parameters to control the random walk process and make it biased random walk.
SDNE~\cite{SDNE} uses deep neural networks to preserve the neighbors structure proximity in network embedding. Recently, graph neural network (GNN) based methods~\cite{SAGE,ArmandpourDHH19,YangPZP0RL20} have been proposed, which generate embedding by sampling and aggregating features from a node local neighborhood in the network. GraphSAGE~\cite{SAGE} is a general inductive framework that leverages node feature information to efficiently generate node embeddings for previously unseen data.
DEGNN~\cite{LiWWL20} presents a distance encoding GNN that learns a generator to approximate the node connectivity distribution and a discriminator to differentiate fake nodes and the nodes sampled from the true data distribution. For a comprehensive review on GNN models, please refer to \cite{WuPCLZY21}.
%A robust negative sampling method~\cite{ArmandpourDHH19} is proposed for network embedding.
\begin{table}
\small
\begin{center}
\begin{tabular}{|c|c|c|c|c|}
\hline
methods & Deep  & Partial  & Multiplex  & Network\\
\hline
DeepWalk~\cite{DeepWalk} &  \textcolor{red}{x}&  \textcolor{red}{x} & \textcolor{red}{x} & \textcolor{green}{\checkmark}\\
Node2Vec~\cite{node2vec} &  \textcolor{red}{x}&  \textcolor{red}{x} & \textcolor{red}{x} & \textcolor{green}{\checkmark}\\
LINE~\cite{LINE} &  \textcolor{red}{x}&  \textcolor{red}{x} &\textcolor{red}{x} & \textcolor{green}{\checkmark}\\
GraphSAGE~\cite{SAGE} &\textcolor{green}{\checkmark}&  \textcolor{red}{x} & \textcolor{red}{x}& \textcolor{green}{\checkmark}\\
SDNE~\cite{SDNE} &\textcolor{green}{\checkmark}&  \textcolor{red}{x} & \textcolor{red}{x}& \textcolor{green}{\checkmark}\\
\hline
DANE~\cite{GaoH18b}  & \textcolor{green}{\checkmark}&  \textcolor{red}{x} & \textcolor{green}{\checkmark}& \textcolor{green}{\checkmark}\\
MVE~\cite{MVE}   & \textcolor{red}{x}&  \textcolor{red}{x} & \textcolor{green}{\checkmark}& \textcolor{green}{\checkmark}\\
MNE~\cite{MNE}  & \textcolor{red}{x}&  \textcolor{red}{x} & \textcolor{green}{\checkmark}& \textcolor{green}{\checkmark}\\
CFANE~\cite{Pan0T0021}   & \textcolor{green}{\checkmark}&  \textcolor{red}{x} & \textcolor{green}{\checkmark}& \textcolor{green}{\checkmark}\\
MAGCN~\cite{ChengWTXG20} & \textcolor{green}{\checkmark}&  \textcolor{red}{x} & \textcolor{green}{\checkmark}& \textcolor{green}{\checkmark}\\
HWNN~\cite{SunY0CCSH21} & \textcolor{green}{\checkmark}&  \textcolor{red}{x} & \textcolor{green}{\checkmark}& \textcolor{green}{\checkmark}\\
\hline
PVC~\cite{LiJZ14} & \textcolor{red}{x} & \textcolor{green}{\checkmark} & \textcolor{green}{\checkmark}& \textcolor{red}{x}\\
IMVC~\cite{LiuZLTZY20}  & \textcolor{red}{x} & \textcolor{green}{\checkmark} & \textcolor{green}{\checkmark}& \textcolor{red}{x}\\
\hline
DPMNE & \textcolor{green}{\checkmark} &  \textcolor{green}{\checkmark}& \textcolor{green}{\checkmark} & \textcolor{green}{\checkmark}\\
\hline
\end{tabular}
\end{center}
\caption{Summary of existing network embedding and partial data representation approaches.
Deep: learning deep representations. Partial: handling partial data. Multiplex: incorporating multiplex data. Network: modeling network information.}\label{compare_methods}
\vspace{-8mm}
\end{table}

\subsection{Multiplex Network Embedding}
Various approaches have been proposed to learn network embedding on multiplex data~\cite{ChangHTQAH15,CenZZYZ019,ChenY0DLZ19,abs-2008-10085,abs-2004-00216,abs-2012-12517,YanZT21,XieOCLXYZ21,qifan2021deep}. Such methods effectively integrate information from individual views while exploiting complementary information supplied by different views. MVE~\cite{MVE} is one of the pioneer work in this category. This work combines the information from multiple sources using a weighted voting scheme. MVN2Vec~\cite{mvn2vec} further studies how varied extent of preservation and collaboration can impact the multiplex embedding learning. DANE~\cite{GaoH18b}, DONE~\cite{BandyopadhyayNV20} and ProGAN~\cite{ProGAN} treat the attribute/tag information associated with the nodes as additional feature and learn embedding with deep neural networks. More recently, MNE~\cite{MNE} uses a latent space to integrate the information across multiple views. MAGCN~\cite{ChengWTXG20} proposes a multi-view attribute graph convolution networks model for the clustering task. A quick survey on multiplex network representation is provided in~\cite{abs-2004-00216}.
These multiplex network embedding methods~\cite{WangJSWYCY19,HuDWS20,XueYRJWL21} achieve promising results. However, the partial data scenarios are not considered.
\begin{figure*}
\begin{center}
\includegraphics[width=1\linewidth]{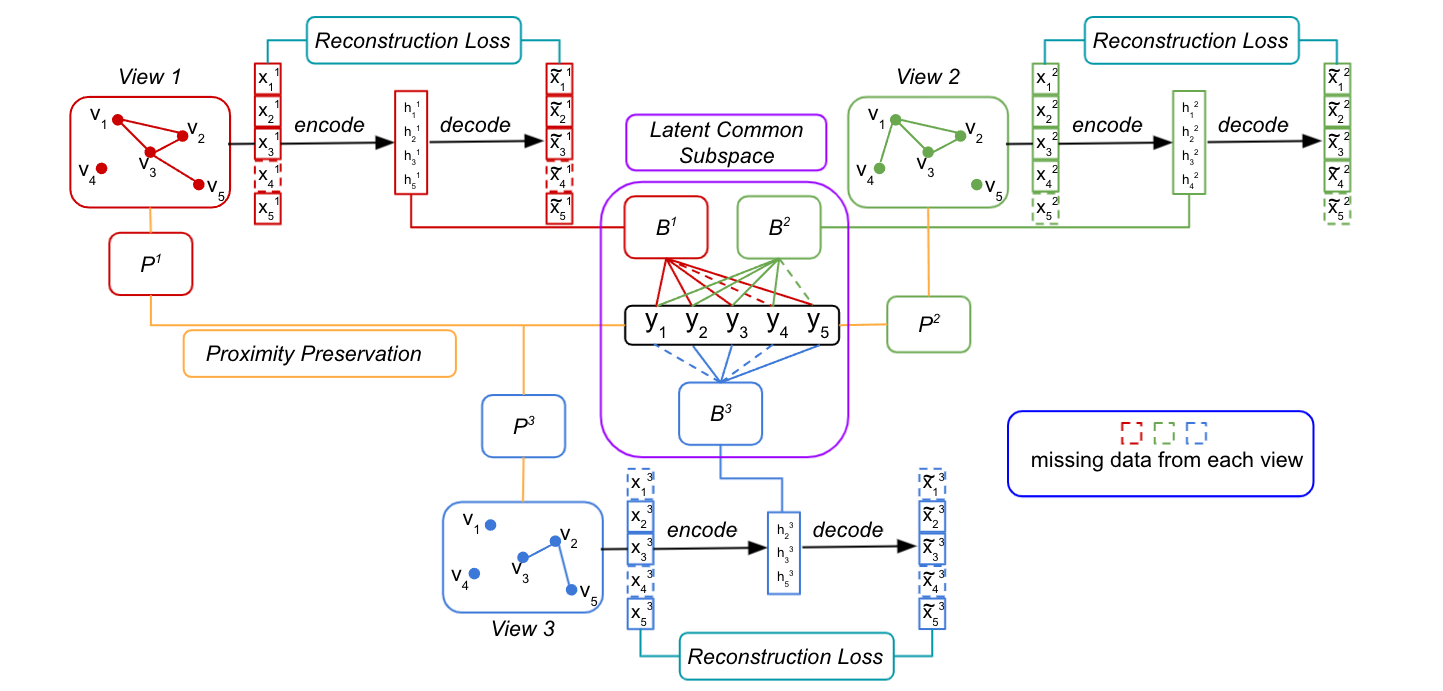}
\end{center}
\caption{The architecture of our proposed DPMNE. There are three views in the multiplex network example, with missing data from each individual views, i.e., $x_4$ is missing from view 1, $x_5$ is missing from view 2, $x_1$ and $x_4$ are missing from view 3. The three key components of DPMNE are: 1) Deep reconstruction loss from all views. 2) Data consistency among views via latent common subspace learning. 3) Proximity preservation from each view.} \label{fig:overview}
\end{figure*}

\subsection{Partial Data Learning}
It is also worth mentioning that there are few single network embedding approaches that deal with incomplete data~\cite{WeiCSLY16,ZhangYZZ18}. Obviously, these methods are not suitable in multiplex network setting. There are also some multi-view approaches~\cite{LiJZ14,RaiNCD16,GuoY19,XuGZWNL19,LiuZLTZY20} proposed to deal with incomplete data in clustering task. For example, PVC~\cite{LiJZ14} develops a two-view clustering method that learns a unique representation between views to handle incomplete data in each view. Most recently, IMVC~\cite{LiuZLTZY20} uses multiple kernel k-means with incomplete kernels to jointly conduct clustering and kernel matrix imputation. Although these methods obtain reasonable performance, they are not directly applicable to network embedding as network information can not be modeled by these methods. Moreover, none of these methods learn the deep representations for the multiplex network. We compare and summarize these approaches in Table \ref{compare_methods}.

\section{Deep Partial Multiplex Network Embedding}
\subsection{Problem Definition}
Given a multiplex network $\pmb{G}$=$(\pmb{V}, \pmb{E})$, where $\pmb{V}$=$\{v_i \ | \ i=1,\dots,n\}$ denotes the set of nodes. $\pmb{X}$=$\{\pmb{X}^1,\pmb{X}^2,\dots,\pmb{X}^t\}$ are the multiplex features associated with the nodes, where $t$ is the total number of views,
$\pmb{X}^s$=\{$x_i^s \ | \ i = 1,2,\dots,n\}$ with $x_i^s\in \mathbb{R}^{d_s}$ is the feature of the $i$-$th$ node in the $s$-$th$ view. $\pmb{E}$=$\{\pmb{E}^{s}\}$ denotes the edge sets in the multiplex network, where $\pmb{E}_{i,j}^s$=$1$ indicates that $v_i$ and $v_j$ are linked in the $s$-$th$ view.
In the partial data setting, a partial data ${\bar{\pmb{X}}}$=$\{{\bar{\pmb{X}}}^{1},{\bar{\pmb{X}}}^{2},\dots,{\bar{\pmb{X}}}^{t}\}$ instead of $\pmb{X}$ is given, where some of the node features are missing from each individual views, e.g., $x_4^1$ is missing from View 1 and $x_5^2$ is missing from View 2 in Figure \ref{fig:overview}. The purpose of DPMNE is to learn a low-dimensional embedding representation $\pmb{Y}$=$\{y_1,y_2,\dots,y_n\}\in \mathbb{R}^{n\times d}$ of $\pmb{G}$, where $d\ll d_s$ is the latent embedding dimension.

The overall model architecture is shown in Figure \ref{fig:overview}. The objective function of DPMNE is composed of three components: (1) Deep reconstruction loss within each views, where we learn a deep representation of the node feature using autoencoder model. (2) Data consistency among views, where latent common subspace learning is utilized to ensure that the node embeddings generated from different views are consistent. (3) Proximity preservation within view, where graph Laplacian is applied to enforce that linked nodes within each network should have close embeddings.

\subsection{Deep Representation}
To capture the sparsity and highly non-linear structure in the feature space, we adopt a deep autoencoder to map the input data to the representation space. Autoencoder is a powerful unsupervised deep model for feature learning. It has been widely
used for various machine learning applications~\cite{JiangGCH16,SDNE}. The basic autoencoder contains three layers, they are
the input layer, the hidden layer, and the output layer, which
are defined as follows:
\begin{equation}\label{eq:01}
h = \sigma(\pmb{W}^{(en)}x+b^{(en)}), \ \ \ \ \tilde{x} = \sigma(\pmb{W}^{(de)}h+b^{(de)})
\end{equation}
where $h$ is the deep representation from the encoder, $\tilde{x}$ is the reconstructed feature from the decoder. $\pmb{H}$ = $\{\pmb{W}^{(en)},b^{(en)},\pmb{W}^{(de)},b^{(de)}\}$ are autoencoder parameters. $\sigma(.)$ denotes the non-linear activation function. In our implementation, we employ $K$ layers in the encoder:
\begin{equation}\label{layers}
\begin{split}
h_1 = \sigma(\pmb{W}_1^{(en)}x+b_1^{(en)}) \ \ \ \ \ \ \\
h_k = \sigma(\pmb{W}_k^{(en)}h_{k-1}+b_k^{(en)})
\end{split}
\end{equation}
Similarly, there will be $K$ layers in the decoder. The goal of the autoencoder is
to minimize the reconstruction loss of the reconstructed features $\tilde{x}$ and the input features $x$, in each individual views (note that there is one autoencoder per view):
\begin{equation}\label{reconstruction_loss}
\min_{\pmb{H}} \sum_{s=1}^t\|{{\pmb{X}}}^{s}-{\tilde{\pmb{X}}}^{s}\|_F^2
\end{equation}
Although minimizing the reconstruction loss does not explicitly preserve the similarity between samples, as shown in~\cite{SalakhutdinovH09}, the reconstruction criterion can effectively capture the data manifolds and thus preserve the similarity between data examples.

\subsection{Data Consistency among Views}
% To capture the sparsity and highly non-linear structure in the feature space, we adopt a deep encoder~\cite{SDNE} to map the input data to the representation space:
% \begin{equation}\label{eq:01}
% \begin{split}
% h_1 = \sigma(\pmb{W}_1x+b_1) \ \ \ \ \ \ \\
% h_k = \sigma(\pmb{W}_kh_{k-1}+b_k)
% \end{split}
% \end{equation}
% where $h_i$ is the deep representation from the encoder. $\pmb{W}$ and $b$ are model parameters. $\sigma(.)$ denotes the non-linear activation function. $k$ is the index of the layers.

In the partial multiplex network setting, the nodes are represented
by heterogeneous features of different dimensions, which makes it difficult for learning their embeddings. By investigating the problem from view perspective, in each individual view, the nodes are sharing the same feature space, and different views are coupled/bridged by the shared common nodes. If we can learn a common latent subspace across views, where embeddings belonging to the same node among different views are consistent, while at the same time for each view, the representations for linked nodes are close in the latent subspace. Then the embeddings can be directly learned from this subspace, and we do not need to fill in or complete the partial data.
%Let ${\hat{\pmb{X}}}^{c}=[{\hat{\pmb{X}}}_c^{1},...,\hat{\pmb{X}}_c^{s},...,{\hat{\pmb{X}}}_c^{t}]$, where ${\hat{\pmb{X}}}_c^{s}\in \mathbb{R}^{c\times d_s}$ are the instances of the common nodes. We denote the instances of each view as ${\bar{\pmb{X}}}^{s}=[{\hat{\pmb{X}}}_c^{s}, {\hat{\pmb{X}}}^{s}]$.
Following the above idea, the deep latent subspace learning can be formulated as:
\begin{equation}\label{eq:02}
\begin{split}
\min_{\pmb{Y},\pmb{B},\pmb{H}} \sum_{s=1}^t\|{\pmb{I}}^{s}({\pmb{H}}^{s}-{\pmb{Y}}{\pmb{B}}^{s})\|_F^2 + \lambda \ R({\pmb{Y}, \pmb{B}, \pmb{H}})
\end{split}
\end{equation}
% \begin{equation}\label{eq:02}
% \begin{split}
% \min_{\pmb{Y},\pmb{B},\pmb{H}} \sum_{s=1}^t\|{\bar{\pmb{H}}}^{s}-{\bar{\pmb{Y}}}^{s}{\pmb{B}}^{s}\|_F^2 + \lambda \ R({\pmb{Y}, \pmb{B}, \pmb{H}})
% \end{split}
% \end{equation}
%\begin{equation}\label{eq:02}
%\begin{split}
%\min_{{\bar{Y}}^{(1)},{B}^{(1)},H^{(1)}} \|H^{(1)}({\bar{X}}^{(1)})-{\bar{Y}}^{(1)}{B}^{(1)}\|_F^2 + \lambda \ R({\bar{Y}}^{(1)},{B}^{(1)})
%\end{split}
%\end{equation}
%\begin{equation}\label{eq:03}
%\begin{split}
%\min_{{\bar{Y}}^{(2)},{B}^{(2)},H^{(2)}} \|{H^{(2)}(\bar{X}}^{(2)})-{\bar{Y}}^{(2)}{B}^{(2)}\|_F^2 + \lambda \ R({\bar{Y}}^{(2)},{B}^{(2)})
%\end{split}
%\end{equation}
where $\pmb{I}^{s}\in \mathbb{R}^{n\times n}$ is a diagonal matrix with element $\pmb{I}^{s}_{ii}$ = 0 or 1 indicating whether $x_i^s$ is missing or not. $\pmb{H}^s\in \mathbb{R}^{n\times d_s^h}$ are the deep representations of the data in $s$-$th$ view as described in Section 3.2. Note that the $i$-$th$ row of $\pmb{H}^s$ will be 0 if $x_i^s$ is missing.
$\pmb{B}^{s}\in \mathbb{R}^{d\times d_s^h}$ is the basis matrix for $s$-$th$ view's latent space. The same latent space dimension $d$ is shared across views.
$\pmb{Y}\in \mathbb{R}^{n\times d}$ is the common representation/embedding of nodes in the latent space.
% $n_s$ ($\le n$) is the number of existing node features in $s$-$th$ view. ${\hat{\pmb{Y}}}_c^{s}$ is latent representation of the common nodes in $s$-$th$ view that share with other views. ${\hat{\pmb{Y}}}^{s}$ represents those nodes only in $s$-$th$ view. Take view 2 in Figure \ref{fig:overview} for example, $n_s$ = 4 (not 5) since there is one missing feature $x_5^2$. ${\hat{\pmb{Y}}}_c^{2}$ = $\{y_1^2, y_2^2, y_3^2\}$ since these three nodes share features from other views, and ${\hat{\pmb{Y}}}^{2}$ = $\{y_4^2\}$ since node $v_4$ only appears in view 2 (missing from both view 1 and view 3).
% $\bar{\pmb{H}}$ are the deep representations of the data described in Section 3.2.
$R(\cdotp)=\|\cdotp\|_F^2$ is the regularization term and $\lambda$ is the trade-off parameter.

%By solving Eqn.\ref{eq:02}, the deep representation $\pmb{H}$, the latent space basis $\pmb{B}$ and corresponding node embedding $\pmb{Y}$ are simultaneously learned to minimize the latent representation error.

% In the above equation, the latent space are learned independently for each view. But in the partial data setting, for nodes present in common views, their embedding ${\hat{\pmb{Y}}}_c^{s}$ should also be consistent. Incorporating the above formulations by ensuring ${\hat{\pmb{Y}}}_c^{s}={\hat{\pmb{Y}}}_c$ for all $s$, we seek to minimize:
% \begin{equation}\label{eq:04}
% \begin{split}
% \min_{\pmb{Y},\pmb{B},\pmb{H}} \sum_{s=1}^t  \left\|\left[
% \begin{array}{ll}
%     {\hat{\pmb{H}}}_c^{s} \\
%     {\hat{\pmb{H}}}^{s} \\
% \end{array}
% \right]-\left[
% \begin{array}{ll}
%     {\hat{\pmb{Y}}}_c \\
%     {\hat{\pmb{Y}}}^{s} \\
% \end{array}
% \right]\pmb{B}^{s}\right\|_F^2 + \lambda \ R(\pmb{Y},\pmb{B},\pmb{H})
% \end{split}
% \end{equation}

In the above formulation, the individual basis matrix $\pmb{B}^{s}$, which are learned from all available instances from all views, are connected by the common latent representation $\pmb{Y}$. Moreover, no interpolation is needed for the missing data beforehand. In fact, the deep representation of the missing data can even be reconstructed with the learned basis and the common latent embedding, i.e., $h_i^s = y_iB^s$.
By solving the above problem, the deep representation $\pmb{H}$, the latent space basis $\pmb{B}$ and the homogeneous feature representation $\pmb{Y}$ are simultaneously learned to minimize the latent representation error.

\subsection{Proximity Preservation within Views}
One of the key problems in network embedding algorithms is proximity preserving, which indicates that linked nodes should be mapped to similar embedding within a close distance. Therefore, besides the data consistency across different views, we also preserve the data proximity within each individual network.
In other words, we want the learned embedding $\pmb{Y}$ to preserve the proximity structure in each network. In this work, we use the $L_2$ distance to measure the proximity between $y_i$ and $y_j$ as $\|y_i-y_j\|^2$ as in most network embedding work. Then one natural way to preserve the proximity in each network is to minimize the weighted average distance as follows:
\begin{equation}\label{eq:05}
\sum_{s=1}^t\sum_{i,j}{\pmb{P}}_{ij}^{s}\|y_i-y_j\|^2 = \sum_{i,j}\pmb{P}_{ij}\|y_i-y_j\|^2
\end{equation}
here $\pmb{P}_{ij}=\sum_{s=1}^t\pmb{P}_{ij}^{s}$ and $\pmb{P}^{s}$ is the proximity matrix in $s$-$th$ view, which can be obtained from the edges in $s$-$th$ network $\pmb{E}^{s}$. A simple way to define $\pmb{P}^{s}$ is to directly use the first-order proximity, i.e., $\pmb{P}^{s}=\pmb{E}^s$. However, the first-order proximity is usually very sparse and insufficient to fully model the relationships between nodes in most cases, especially under the partial data setting. In order to characterize the connections between nodes better, we adopt high-order proximity~\cite{GaoH18b,Li0ZZC19} and define $\pmb{P}^s$ as:
\begin{equation}\label{high-order}
\pmb{P}^s = w_1\pmb{E}^s + w_2(\pmb{E}^s)^2 + \dots + w_l(\pmb{E}^s)^l
\end{equation}
where $l$ is the order, and $w_1$,$\dots$,$w_l$ are the weights for each term\footnote{In our implementation, we set $l$ to 5, $w_1$ to 1 and $w_i$ = $0.5w_{i-1}$.}. Matrix $(\pmb{E}^s)^l$ denotes the $l$-order proximity matrix. To meet the proximity preservation criterion, we seek to minimize the quantity in Eqn.\ref{eq:05} in the network since it incurs a heavy penalty if two connected nodes have very different embedding representations. By introducing a diagonal matrix $\pmb{D}$, whose entries are given by $\pmb{D}_{ii}=\sum_{j=1}^n \pmb{P}_{ij}$.
Eqn.\ref{eq:05} can be rewritten as:
\begin{equation}\label{eq:06}
\begin{split}
tr\left({\pmb{Y}}^T(\pmb{D}-\pmb{P}){\pmb{Y}}\right)=tr\left(\pmb{Y}^T\pmb{L}{\pmb{Y}}\right)
\end{split}
\end{equation}
where $\pmb{L}$ is called graph $Laplacian$~\cite{BelkinN01} and $tr(\cdotp)$ is the matrix trace function.
By minimizing the above objective in all networks, the proximity between different nodes can be preserved in the learned embedding.

\subsection{Overall Objective and Optimization}
The entire objective function consists of three components: the deep reconstruction loss in Eqn.\ref{reconstruction_loss}, the data consistency among views in Eqn.\ref{eq:02} and proximity preservation within views given in Eqn.\ref{eq:06} as follows:
\begin{equation}\label{eq:07}
\begin{split}
\min_{\pmb{Y},\pmb{B},\pmb{H}} O=\sum_{s=1}^t\|{{\pmb{X}}}^{s}-{\tilde{\pmb{X}}}^{s}\|_F^2 + \alpha \sum_{s=1}^t \|{\pmb{I}}^{s}({\pmb{H}}^{s}-{\pmb{Y}}{\pmb{B}}^{s})\|_F^2\\
+ \beta \ tr\left(\pmb{Y}^T\pmb{L}{\pmb{Y}}\right) + \lambda \ R(\pmb{Y},\pmb{B},\pmb{H}) \ \ \ \ \ \ \ \ \ \ \ \
\end{split}
\end{equation}
where $\alpha$, $\beta$ and $\lambda$ are trade-off parameters to balance the weights among the terms.
%Note that ${\bar{\pmb{Y}}}^{s}$ share an identical part ${\hat{\pmb{Y}}}_c$ corresponding to the common nodes present across views.
Directly minimizing the objective function in Eqn.\ref{eq:07} is intractable since it is a non-convex optimization problem with $\pmb{Y}$, $\pmb{B}$ and $\pmb{H}$ coupled together. We propose to use coordinate descent scheme by iteratively solving the optimization problem with respect to $\pmb{Y}$, $\pmb{B}$ and $\pmb{H}$ as follows:

\textbf{(1) Update ${Y}$ by fixing $B$ and $H$}. Given the basis matrix $\pmb{B}^{s}$ and encoders $\pmb{H}^{s}$ for all views, we seek to solve the following sub-problem:
\begin{equation}\label{eq:08}
\begin{split}
\min_{\pmb{Y}} \ O(\pmb{Y}) \ = \alpha \sum_{s=1}^t \|{\pmb{I}}^{s}({\pmb{H}}^{s}-{\pmb{Y}}{\pmb{B}}^{s})\|_F^2\\ \ \ \ \ \ \ \ \ \ \ \ \ \ \ \ \ \ \ \ \ \ \ \\
+ \beta \ tr\left(\pmb{Y}^T\pmb{L}{\pmb{Y}}\right) + \lambda \ R({\pmb{Y}})  + const
\end{split}
\end{equation}
where $const$ is the constant value independent with the parameter that to be optimized with. Although Eqn.\ref{eq:08} is still non-convex, it is smooth and differentiable which enables gradient descent methods for efficient optimization with the calculated gradient:
\begin{equation}\label{eq:09}
\begin{split}
\partial\frac{O(\pmb{Y})}{\pmb{Y}}=2{\pmb{I}}^{s}(\pmb{Y}\pmb{B}^{s}-{\pmb{H}}^{s})(\pmb{B}^{s})^T+2\beta\pmb{L}{\pmb{Y}}+4\lambda\pmb{Y}(\pmb{Y}^T\pmb{Y}-\pmb{I}_d)
\end{split}
\end{equation}

\textbf{(2) Update ${B}^{s}$ by fixing $Y$ and $H$}. It is equivalent to solve the following least square problems:
\begin{equation}\label{eq:12}
\min_{\pmb{B}^{s}} O(\pmb{B}^{s}) = \alpha \|{\pmb{I}}^{s}({\pmb{H}}^{s}-{\pmb{Y}}{\pmb{B}}^{s})\|_F^2 + \lambda \|\pmb{B}^{s}\|_F^2
\end{equation}
which has a closed form solution and can be simply derived.

\textbf{(3) Update ${H}^{s}$ by fixing $Y$ and $B$}. It is a standard autoencoder with an additional regression loss:
\begin{equation}\label{eq:13}
\min_{\pmb{H}^{s}} O(\pmb{H}^{s}) = \|{{\pmb{X}}}^{s}-{\tilde{\pmb{X}}}^{s}\|_F^2 + \alpha \|{\pmb{I}}^{s}({\pmb{H}}^{s}-{\pmb{Y}}{\pmb{B}}^{s})\|_F^2  + \lambda \ R(\pmb{H}^{s})
\end{equation}
which can be solved with gradient back-propagation.
We then alternate the process of updating $\pmb{Y}$, $\pmb{B}$ and $\pmb{H}$ for several iterations to find a locally optimal solution.
% The full learning algorithm is described in Algorithm 1.
% \begin{algorithm}
% \small
% \caption{Deep Partial Multiplex Network Embedding (DPMNE)}
% \begin{algorithmic}
% \Require
% Partial data ${\bar{\pmb{X}}}$=$\{{\bar{\pmb{X}}}^{1},{\bar{\pmb{X}}}^{2},\dots,{\bar{\pmb{X}}}^{t}\}$, edge sets $\pmb{E}^{s}$, trade-off parameters $\alpha$, $\beta$ and $\lambda$
% \Ensure
% Unified node embeddings $\pmb{Y}$.
% \State Initialize the basis matrices $\pmb{B}$ and the autoencoder $\pmb{H}$, Calculate the graph Laplacian $\pmb{L}$.
% \Repeat
% \State Optimize Eqns.\ref{eq:08} with gradient in Eqns.\ref{eq:09} and update embeddings $\pmb{Y}$.
% \State Optimize Eqn.\ref{eq:12} and update the basis matrix $\pmb{B}^{s}$.
% \State Optimize Eqn.\ref{eq:13} and update the autoencoder $\pmb{H}^{s}$.
% \Until the solution converges
% \end{algorithmic}
% \end{algorithm}

\subsection{Binary Embedding}
This section connects our work to the quantization-based binary embedding techniques~\cite{GongLGP13,WangZSSS18,wang2015learning}, which learn compact binary representations of the data examples for efficient similarity search tasks.
% We first state and prove the following orthogonal invariant property of our learned embeddings:
% \begin{theorem}
% Assume $\pmb{Q}$ is a $d\times d$ orthogonal matrix, i.e., $\pmb{Q}^T\pmb{Q}={\pmb{I}_d}$. If $\pmb{Y}$, $\pmb{B}$ and $\pmb{H}$ are an optimal solution to the problem in Eqn.\ref{eq:07}, then $\pmb{Y}^{'}$=$\pmb{YQ}$, $\pmb{B}^{'}$=$\pmb{Q}^T\pmb{B}$ and $\pmb{H}^{'}$=$\pmb{H}$ are also an optimal solution.
% \end{theorem}
% \begin{proof}
% By substituting $\pmb{YQ}$ and $\pmb{Q}^T\pmb{B}$ into Equation \ref{eq:07}, it is obvious that:
% $\|\pmb{I}^{s}({\pmb{H}}^{s}-{{\pmb{Y}}}\pmb{Q}\pmb{Q}^T\pmb{B}^{s})\|_F^2=\|{\pmb{I}}^{s}({\pmb{H}}^{s}-{\pmb{Y}}{\pmb{B}}^{s})\|_F^2$,
% $tr\left({({\pmb{Y}}\pmb{Q})}^T\pmb{L}{{{\pmb{Y}}}\pmb{Q}}\right)=tr\left(\pmb{Q}^T\pmb{Y}^T\pmb{L}{{{\pmb{Y}}}\pmb{Q}}\right)=tr\left(\pmb{Y}^T\pmb{L}\pmb{Y}\right)$,
% and $\|\pmb{YQ}\|_F^2=\|\pmb{Y}\|_F^2$, $\|\pmb{Q}^T\pmb{B}\|_F^2=\|\pmb{B}\|_F^2$. Thus, the value of the objective function in the equation does not change by the orthogonal rotation.
% \end{proof}
Quantization-based binary embedding methods directly binarize the low-dimensional representation to achieve the binary codes. In this work, we can easily obtain the binary codes $\pmb{C}$ for the nodes in the network by binarizing the learned embedding $\pmb{Y}$. However, the quantization error can be further reduced based on the orthogonal invariant property, by minimizing the quantization error between the binary codes and the orthogonal rotation of the embeddings (since $\pmb{YQ}$ is also an optimal embedding):
\begin{equation}\label{eq:14}
\begin{split}
\min_{{\pmb{C},\pmb{Q}}} \|{\pmb{C}}-\pmb{YQ}\|_F^2 \ \ \ \ \ \ \ \ \ \ \ \ \ \ \ \ \ \ \ \ \ \ \ \\
s.t. \ \ \ \ \ \ \pmb{C}\in \{-1,1\}^{n\times d}, \ \ \ \pmb{Q}^T\pmb{Q}={\pmb{I}_d} \ \ \ \ \ \
\end{split}
\end{equation}
The above quantization problem is well studied in the literature~\cite{GongLGP13}.

\subsection{Theoretical Analysis}
This section provides some complexity analysis on the training cost of the learning algorithm.
The optimization algorithm of DPMNE consists of three steps in each iteration to update $Y$, $B$ and $H$. The time complexities for solving ${Y}$ and $B$ are bounded by $O(tndd_s+tnd^2+n^2d)$ and $O(tnd^2+tndd_s)$ respectively. In practice, $L$ is usually a sparse matrix, and the cost can be reduced from $O(n^2d)$ to $O(ld)$ with sparse matrix multiplication, where $l$ is the number of non-zero elements in $L$. The cost of updating $H$ depends on the number of hidden layers and units in the autoencoder network, which is roughly $O(tnmd_s)$. Here $m$ is the number of unique units in the network. Thus, the total time complexity of the learning algorithm is bounded by $O(tndd_s+ld+tnd^2+tnmd_s)$.
\begin{table}
\small
\begin{center}
%resizebox{0.45\textwidth}{!}{
\begin{tabular}{|c|c|c|c|c|c|}
\hline
Dataset & \#nodes  & \#total edges  & \#views  & \#labels & avg. PDR\\
\hline
{\bf{Cora}} & 2,708 &  12,887& 2& 7& 0.02\\
\hline
{\bf{DBLP}} & 69,110 &  1,884,236& 3& 8& 0.39\\
\hline
{\bf{Flickr}} & 6,163 &  378,547& 5& 10& 0.46\\
\hline
{\bf{Last.fm}} & 10,197 &  1,325,367 & 12& 11& 0.52\\
\hline
\end{tabular}
%}
\end{center}
\caption{A summary of the statistics on all datasets.}\label{table:data}
\vspace{-8mm}
\end{table}

\begin{table*}
\small
\begin{center}
\resizebox{1.0\textwidth}{!}{
\begin{tabular}{|c|c|c|c|c|c|c|c|c|}
\hline
 &\multicolumn{2}{|c|}{Cora} &\multicolumn{2}{|c|}{DBLP} &\multicolumn{2}{|c|}{Flickr}&\multicolumn{2}{|c|}{Last.fm}\\
\hline
 methods &  Micro-F1 & Macro-F1 &  Micro-F1 & Macro-F1 & Micro-F1 & Macro-F1 & Micro-F1 & Macro-F1 \\
\hline
DeepWalk & 0.817 $\pm$ 0.022 & 0.809 $\pm$ 0.019 & 0.709 $\pm$ 0.014 & 0.713 $\pm$ 0.010 & 0.472 $\pm$ 0.015 & 0.451 $\pm$ 0.013 & 0.482 $\pm$ 0.011 & 0.453 $\pm$ 0.014\\
GraphSAGE & 0.826 $\pm$ 0.015 & 0.814 $\pm$ 0.017 & 0.718 $\pm$ 0.019 & 0.720 $\pm$ 0.021 & 0.479 $\pm$ 0.015 & 0.462 $\pm$ 0.014 & 0.503 $\pm$ 0.011 & 0.465 $\pm$ 0.012\\
SDNE & 0.821 $\pm$ 0.021 & 0.816 $\pm$ 0.016 & 0.706 $\pm$ 0.014 & 0.709 $\pm$ 0.018 & 0.482 $\pm$ 0.015 & 0.454 $\pm$ 0.009 & 0.508 $\pm$ 0.011 & 0.479 $\pm$ 0.019\\
\hline
DANE & 0.841 $\pm$ 0.015 & 0.832 $\pm$ 0.012 & 0.727 $\pm$ 0.015 & 0.721 $\pm$ 0.014 & 0.485 $\pm$ 0.021 & 0.467 $\pm$ 0.013 & 0.503 $\pm$ 0.014 & 0.481 $\pm$ 0.012\\
%MEGAN & 0.844 $\pm$ 0.031 & 0.835 $\pm$ 0.024 & 0.741 $\pm$ 0.022 & 0.725 $\pm$ 0.023 & 0.506 $\pm$ 0.019 & 0.483 $\pm$ 0.016 & 0.524 $\pm$ 0.012 & 0.507 $\pm$ 0.007\\
IMVC & 0.828 $\pm$ 0.017 & 0.820 $\pm$ 0.016 & 0.741 $\pm$ 0.013 & 0.734 $\pm$ 0.016 & 0.478 $\pm$ 0.020 & 0.464 $\pm$ 0.012 & 0.496 $\pm$ 0.017 & 0.478 $\pm$ 0.013\\
MAGCN & 0.856 $\pm$ 0.022 & \textbf{0.847 $\pm$ 0.021} & 0.750 $\pm$ 0.025 & 0.736 $\pm$ 0.019 & 0.508 $\pm$ 0.018 & 0.480 $\pm$ 0.015 & 0.530 $\pm$ 0.011 & 0.508 $\pm$ 0.012\\
HWNN & 0.851 $\pm$ 0.017 & 0.843 $\pm$ 0.016 & 0.752 $\pm$ 0.023 & 0.741 $\pm$ 0.014 & 0.502 $\pm$ 0.013 & 0.474 $\pm$ 0.021 & 0.532 $\pm$ 0.013 & 0.511 $\pm$ 0.016\\
\hline
DPMNE & \textbf{0.859 $\pm$ 0.016} & 0.844 $\pm$ 0.015 & \textbf{0.784 $\pm$ 0.016} & \textbf{0.769 $\pm$ 0.009} & \textbf{0.526 $\pm$ 0.011} & \textbf{0.512 $\pm$ 0.014} & \textbf{0.558 $\pm$ 0.013} & \textbf{0.534 $\pm$ 0.015}\\
\hline
\end{tabular}
}
\end{center}
\caption{Node classification results on all datasets. Results are statistically significant with p-value $<$ 0.001.}\label{t1}
\end{table*}
\section{Experiments}
\subsection{Experimental Setting}
\subsubsection{Datasets}
The proposed approach is evaluated on four benchmarks:  {\bf{Cora}}, {\bf{DBLP}}, {\bf{Flickr}} and {\bf{Last.fm}}.
\begin{itemize}
\item {\bf{Cora}}\footnote{\url{https://linqs-data.soe.ucsc.edu/public/lbc/cora.tgz}} is a widely used document corpus from paper citation networks. It contains 2,708 scientific publications classified into one of 7 classes. Citation links and attributes are used as the multiplex, where the features are represented by a 0/1-valued vector indicating the absence/presence of the corresponding link and attribute respectively. The citation network consists of 5,429 edges. For the attribute network, there is an edge between two papers if they share at least one attribute, resulting in 7,458 edges.
\item {\bf{DBLP}}\footnote{\url{https://www.aminer.org/aminernetwork}} is an author network from the DBLP dataset~\cite{TangZYLZS08}. It contains 69,110 nodes.
Three views are identified including the co-authorship, author-citation and text views. For co-authorship and author-citation views, 0/1-valued feature vectors are used indicating co-authorship and citation between two authors, resulting in 430,117 and 763,029 edges in the corresponding views. For the text view, TF-IDF features are extracted from author's title and abstract. The network is construct based on the text similarity, i.e., there is a link between two authors if their text similarity is high, resulting in 691,090 edges.
We select eight diverse research fields as labels including ``machine learning'', ``computational linguistics'', ``programming language'', ``data mining'', ``database'', ``system technology'', ``hardware'' and ``theory''. For each field, several representative conferences are
selected, and only papers published in these conferences are kept to construct the three views.
\item {\bf{Flickr}}~\cite{FlickrData} data were collected from the Flickr photo sharing service. It consists of 6,163 users with 10 unique labels. There are 5 views associated with this data: Comment, Favorite, Photo, Tag and User.
Here, the views correspond to different aspects of Flickr and edges denote shared interests between users. For example, in the
comment view, there is a link between 2 users if they have both commented on the same set of 5 or more photos. All features are 0/1-valued vectors.
The resulting five views are: CommentView (2,358 nodes, 13,789 links), FavoriteView (2,724 nodes, 30,757 links), PhotoView (4,061 nodes, 91,329 links), TagView (1,341 nodes, 154,620 links), and UserView (6,163 nodes, 88,052 links).
\item {\bf{Last.fm}}~\cite{FlickrData} dataset was collected from the music network, with the nodes representing the users and the edges corresponding to different relationships between Last.fm users and other entities. In each view,
two users are connected by an edge if they share similar interests, yielding 12 views: ArtistView (2,118 nodes, 149,495 links), EventView (7,240 nodes, 177,000 links), NeighborView (5,320 nodes, 8,387 links), ShoutView (7,488 nodes, 14,486 links), ReleaseView (4,132 nodes, 129,167 links), TagView (1,024 nodes, 118,770 links), TopAlbumView (4,122 nodes, 128,865 links), TopArtistView (6,436 nodes, 12,4731 links), TopTagView (1,296 nodes, 136,104 links), TopTrackView (6,164 nodes, 87,491 links), TrackView (2,680 nodes, 93,358 links), and UserView (10,197 nodes, 38,743 links).
\end{itemize}
All these multiplex network are suffering from missing data. For example, the CommentView of {\bf{Flickr}} only has 2,358 users (out of 6,163), where comments are missing from a certain amount of users. In the TagView of {\bf{Last.fm}}, only 1,024 out of 10,197 users have associated with tags, resulting partial multiplex data. We use Partial Data Ratio (PDR) to represent the fraction of the missing data, e.g., the PDR of the CommentView of {\bf{Flickr}} is 0.62\footnote{(6,163-2,358)/6,163 = 0.62}. The statistics of the datasets with average PDR are given in Table \ref{table:data}.
\begin{table}
\small
\begin{center}
\begin{tabular}{|c|c|c|c|c|}
\hline
 &{Cora}  &{DBLP}  &{Flickr} &{Last.fm}\\
\hline
DeepWalk &  0.676&  0.573 & 0.429 & 0.435\\
GraphSAGE & 0.683&  0.589 & 0.422 & 0.441\\
SDNE & 0.677 &  0.594& 0.397 & 0.430\\
\hline
DANE  & 0.695&  0.613 & 0.454 & 0.444\\
%MEGAN & 0.701 &  0.622 & 0.468 & 0.456\\
IMVC  & 0.692&  0.635 & 0.457 & 0.447\\
MAGCN & 0.707 &  0.628 & 0.480 & 0.461\\
HWNN & 0.702 &  0.631 & 0.482 & 0.465\\
\hline
DPMNE &\textbf{0.711} &  \textbf{0.649} & \textbf{0.506}& \textbf{0.483}\\
\hline
\end{tabular}
\end{center}
\caption{Node clustering accuracy results on all datasets.}\label{t2}
\end{table}
\begin{figure*}
\begin{center}
\includegraphics[width=1\linewidth]{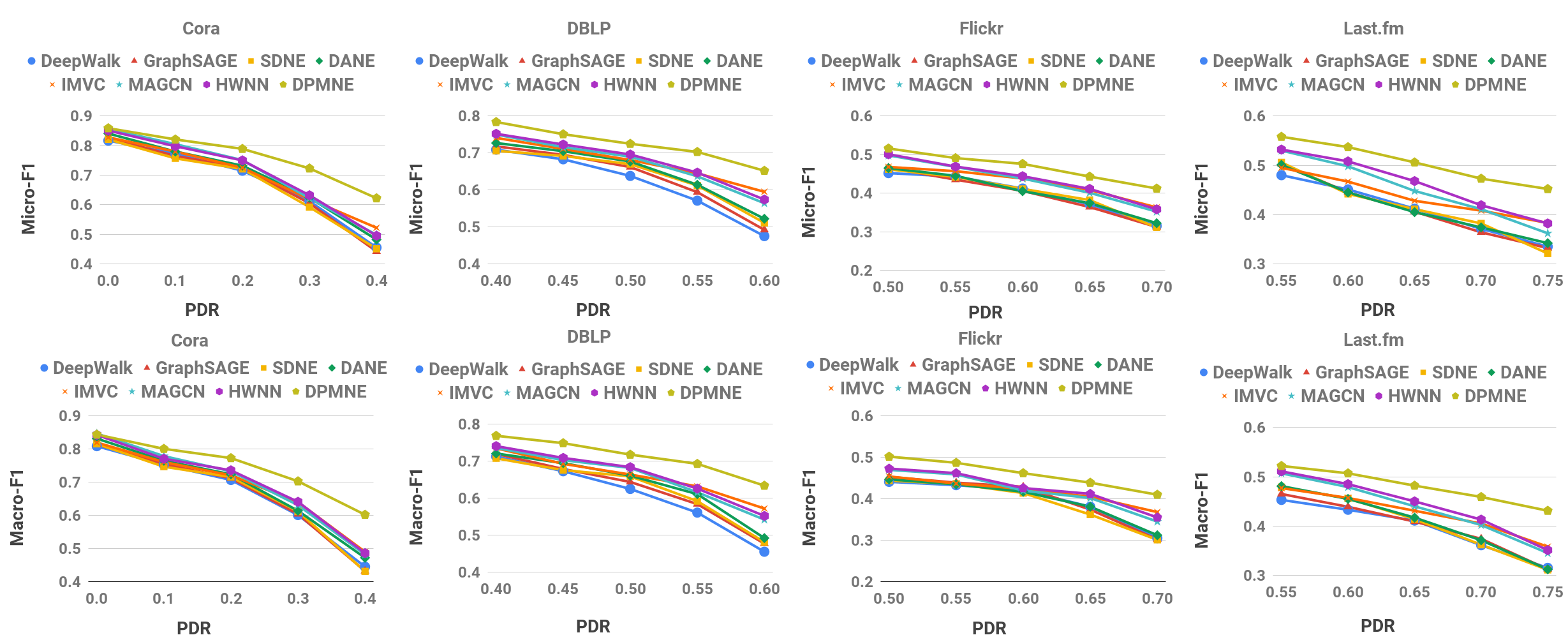}
\end{center}
\caption{Performance of node classification under different PDRs on all datasets.} \label{fig:1}
\end{figure*}

\subsubsection{Baselines}
The proposed approach is compared with seven different state-of-the-art baselines, including three single-view methods, {\bf{DeepWalk}}, {\bf{GraphSAGE}} and {\bf{SDNE}} with four multi-view methods, {\bf{DANE}}, {\bf{IMVC}}, {\bf{MAGCN}} and {\bf{HWNN}}.
%Note that we do not include {\bf{MNE}}~\cite{MNE} and {\bf{MVE}}~\cite{MVE} for comparison as {\bf{MEGAN}} already demonstrates superior performance than them.
\begin{itemize}
\item {\bf{DeepWalk}}~\cite{DeepWalk} learns node representations from random-walk on the networks.
\item {\bf{GraphSAGE}}~\cite{SAGE} aggregates the neighbor information to generate node embeddings with graph neural network.
\item {\bf{SDNE}}~\cite{SDNE} uses deep neural networks to preserve the structure proximity in network embedding.
\item {\bf{DANE}}~\cite{GaoH18b} uses attribute/tag information associated with the nodes as additional feature and learn embedding with deep neural networks.
%\item {\bf{MEGAN}}~\cite{MEGAN} employs a generator that effectively integrates information about pairwise links between nodes.
\item {\bf{IMVC}}~\cite{LiuZLTZY20} is a multi-view clustering method that explicitly handles incomplete data.
\item {\bf{MAGCN}}~\cite{ChengWTXG20} learns network embeddings with a multi-view graph convolution networks (GCN).
\item {\bf{HWNN}}~\cite{SunY0CCSH21} is a GNN-based representation learning for heterogeneous hypergraphs, which models multiple non-pairwise relations.
\end{itemize}

\subsubsection{Implementation Details}
To apply single-view method, we generate a single network from the multiplex network by placing an edge between a pair of nodes if they are linked by an edge in any view. For DeepWalk, we set the window size as 10, the walk length as 80, and the number of walks as 10 which are the optimal parameters tuned in our experiments. The number of units in the hidden layer is set to 200 in the deep neural network for GraphSAGE, SDNE and DANE. For IMVC, the code is public available\footnote{\url{https://github.com/xinwangliu/TPAMI_EEIMVC}} and we tune the best hyperparameter with 5-fold cross validation. %For MAGCN and HWNN, we obtain the codes from the authors and tune for the optimal hyperparameters.

For our method, the parameters $\alpha$, $\beta$ and $\lambda$ are tuned by 5-fold cross validation on the training set. To get a fair comparison with deep models, we adopt the same architecture of the neural network, with 200 units in the hidden layer. We set the maximum number of iterations to 60. The number of embedding dimension is set to 128 for all methods (for DANE, each view has dimension of 64). We repeat each experiment 10 times and report the result based on the average over these runs. 50\% of the data with random split is used as training.

\subsection{Results and Discussion}
\subsubsection{{\bf Evaluation of Different Methods}}
We first evaluate the performance of different methods. To apply the compared deep neural network methods to the partial data, a simple way is to fill in the missing features with 0. However, this may result in large fitting errors across views for the multi-view methods, since the embedding for the missing instance will be 0. Therefore, to achieve stronger baseline results, we replace the missing feature using the linear combination of its 5-nearest neighbor examples, weighted by the similarities, which appear across views. Then the baseline deep methods can be directly applied on these extended data.

We conduct both node classification and clustering on the learned node embedding. Specifically, for classification, we employ L2-regularized logistic regression as the classifier, with Micro-F1 and Macro-F1 as metrics. For clustering, we employ $K$-means as the clustering method and use clustering accuracy as the metric. The classification results with standard deviations on all datasets are reported in Table \ref{t1}. From these comparison results, we can see that DPMNE provides the best results among all compared methods on all datasets. For example, the Micro-F1 of DPMNE increases over 4.3\% and 7.8\% compared with both HWNN and DANE on {\bf{DBLP}}. The reason is that DPMNE can effectively handle the partial data by common latent subspace learning across views and proximity preservation within individual networks, while the compared methods fail to accurately extract a common space from the partial nodes. We observe that DPMNE outperforms IMVC by 8.8\%. Although IMVC tries to deal with incomplete data, the network information is not fully explored.
We also observe that DPMNE achieves comparable or slightly better results compare with other baselines on {\bf{Cora}}, whose PDR is very small, i.e., 0.02 from Table \ref{table:data}. This further validates that DPMNE is equally effective on multiplex network without missing data.
It can be seen that multi-view methods outperform the single-view methods on all four datasets. The reason is that multi-view methods construct embedding that incorporates complementary information from all views. The clustering result is summarized in Table \ref{t2}. From this table, we can find that our approach achieves much better clustering performance than the others for most cases, which further verifies the effectiveness of DPMNE.
\begin{table}
\small
\begin{center}
%\resizebox{0.35\textwidth}{!}{
\begin{tabular}{|c|c|c|c|c|}
\hline
Micro-F1 &{Cora}  &{DBLP}  &{Flickr} &{Last.fm}\\
\hline
w/o deep autoencoder &  0.835&  0.747 & 0.479 & 0.524\\
\hline
w/o partial multiplex & 0.852 &  0.739 & 0.488 & 0.517\\
\hline
w/o proximity preservation & 0.830 &  0.721& 0.477 & 0.535\\
\hline
DPMNE &\textbf{0.859} &  \textbf{0.784} & \textbf{0.526}& \textbf{0.558}\\
\hline
\end{tabular}
%}
\end{center}
\caption{Performance of node classification with different model ablations.}\label{ablation}
\end{table}

\subsubsection{{\bf Effect of Partial Data Ratio}}
To evaluate the effectiveness of the proposed DPMNE under different PDRs, we progressively increase the PDR by randomly removing features from the multiplex network, and compare our method with the other baselines. The node classification results are shown in Figure \ref{fig:1}. It can be seen from the figure that when the partial data ratio is 0 (on {\bf{Cora}}), the data actually becomes the traditional multiplex setting without missing data. As aforementioned, DPMNE is also comparable with other baselines. However, as the PDR increases, our DPMNE approach significantly outperforms other baselines on all datasets. In other words, the performance of DPMNE drops much slower compared to the baseline methods. For example, the Micro-F1 of DPMNE increases over 20\% compared with the state-of-the-art GNN-based models, HWNN and MAGCN, on {\bf{Cora}} with 0.4 PDR.
Our hypothesis is that, although the missing data are recovered from the common nodes across views, the baseline deep methods seem less effective in the view missing case. The missing data may not be accurately recovered when the data are missing blockwise for the partial data setting. In other words, the missing nodes can be dissimilar to all the nodes appear across views.
%Another interesting observation is that the results of single-view methods SDNE and Node2Vec are comparable with multi-view methods with PDR 0 on {\bf{Flickr}} dataset. This is attributed to the fairly rich link information within the Flickr tag view. In this case, single view methods may be competitive with multi-view methods when almost all of the information needed is available.
\begin{figure}
\includegraphics[width=0.9\linewidth]{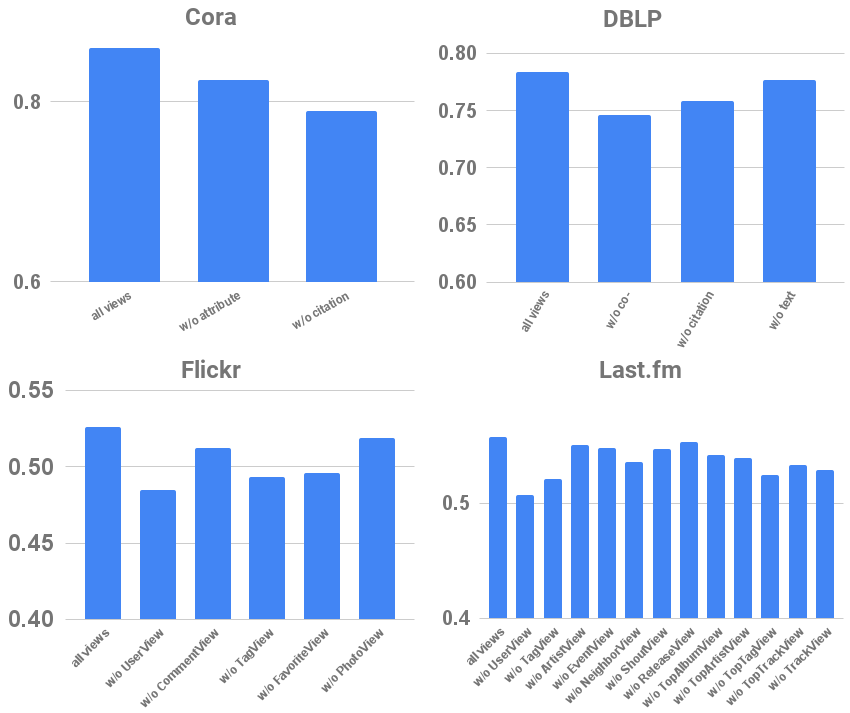}
\caption{View importance chart on all datasets.} \label{fig:view_importance}
\end{figure}

\subsubsection{{\bf Ablation Study}}
%each component
We conduct a series of ablation studies of our model. We first analyze the behavior of each component in DPMNE. There are three key components in our DPMNE, the deep autoencoder, the common latent subspace learning and the network proximity preservation. We train three additional models by removing each of these components separately. Specifically, we remove the deep autoencoder by fixing all its parameters to 1, i.e., both encoder and decoder will map the input to itself.
For the other two components, we simply remove the loss terms in the objective. The comparison results are shown in Table \ref{ablation}.
It can be seen that all these components are indispensable in DPMNE. Without the common latent subspace learning part, our model degrades to HWNN or MAGCN which is not able to model the partial multiplex data effectively. On the other hand, we observe that deep autoencoder clearly improve the model quality. The reason is that it captures the sparsity and non-linearity in the original feature space. Lastly, it is obvious that network proximity preservation is crucial in learning network embeddings.

%each view importance
To understand which views are important for learning the network embeddings, we conduct another set of experiments on view importance analysis. Specifically, we build and evaluate the model performance by
removing one view from the multiplex network at a time. The node classification results are shown in Figure \ref{fig:view_importance}. It can be seen that co-authorship and author-citation views are more important than the text view in {\bf{DBLP}} dataset, which is consistent with our expectation, as text similarity might not truly reflect the relationships among authors. We also observe that the UserView is the most important view in both {\bf{Flickr}} and {\bf{Last.fm}} datasets. The reason is that the UserView directly reveals the connections among the users. Some other useful views are TagView and FavoriteView.

\begin{figure}
\includegraphics[width=0.9\linewidth]{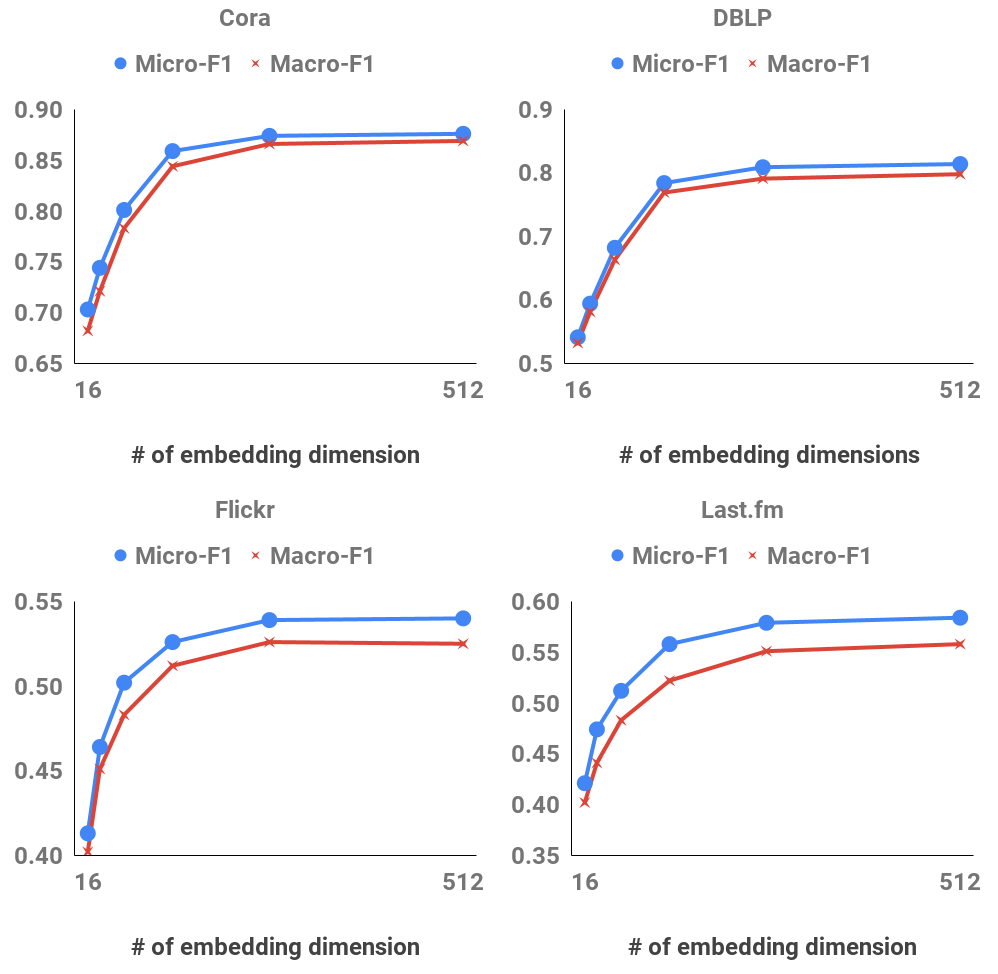}
\caption{Performance of node classification with different numbers of embedding dimension on all datasets.} \label{fig:2}
\end{figure}

%Effect of Embedding Dimension
We evaluate the performance of DPMNE with different embedding dimensions by varying the dimension $d$ from \{16, 32, 64, 128, 256, 512\}. The node classification results on all datasets are shown in Fig.\ref{fig:2}. It can be seen from the figure that the values of both Micro-F1 and Macro-F1 of DPMNE consistently increase with the increasing of the embedding dimension, from 16 to 256, on all datasets. Our approach achieves similar results between dimension 256 and 512. This observation indicates that very large embedding size is not needed for node representation, which is consistent with the observation in MEGAN \cite{MEGAN}.
\subsubsection{{\bf Effect of Binary Embedding}}
We further evaluate our approach in learning binary embeddings. The binary embeddings can be direct achieved by binarizing the learned embeddings, and this method is referred to bi-DPMNE. To obtain more effective binary embeddings, we further conduct iterative quantization based on the orthogonal invariant property of the learned embeddings from Eqn.\ref{eq:14}. This method is referred to bi-DPMNE-ITQ. We then compare these two methods with two state-of-the-art multiplex binary embedding methods, CCA-ITQ~\cite{GongLGP13,GongKIL14} and DCMH~\cite{JiangL17}. The Micro-F1 results with 128 bits are reported in Table \ref{t3}. It can be seen from the table that bi-DPMNE consistently performs better than the two baselines on the incomplete data. On the other hand, the bi-DPMNE-ITQ achieves even better results compared to bi-DPMNE, which is consistent with our expectation as it further minimizes the quantization errors.
\begin{table}
\small
\begin{center}
%\resizebox{0.4\textwidth}{!}{
\begin{tabular}{|c|c|c|c|c|}
\hline
 Micro-F1&{Cora} &{DBLP} &{Flickr}&{Last.fm}\\
\hline
CCA-ITQ &  0.748&  0.657 & 0.438 & 0.451\\
\hline
DCMH  &  0.762&  0.683 & 0.449& 0.474 \\
\hline
bi-DPMNE &  0.788&  0.709 & 0.463 & 0.482\\
bi-DPMNE-ITQ &  0.803 &  {0.727} & {0.472}& {0.495} \\
\hline
\end{tabular}
%}
\end{center}
\caption{Node classification comparison of binary embedding with 128 bits on all datasets.}\label{t3}
\end{table}

% \subsubsection{{\bf Additional Experiments}}
% We conduct parameter sensitivity experiments to demonstrate that DPMNE is relatively stable with respect to the hyper-parameters. Due to the space limitation, we report these results with the link prediction and training cost experiments in the appendix.

\section{Conclusions}
% Network  embedding  is  designed  for  learning  low-dimensional and typically non-linear representations of nodes in the network, which is able to preserve network information. Network embedding has been shown to be useful in many downstream tasks.
% Real-world networks are usually with multiplex or having multi-view representations from different relations.
In this paper, we propose a novel deep network embedding approach to deal with partial multiplex data. We formulate a unified learning framework by simultaneously minimizing the deep reconstruction loss with the autoencoder neural network, ensuring data consistency among different views via common latent subspace learning, and preserving data proximity within the same view through graph Laplacian. Extensive experiments on four benchmarks have demonstrated the effectiveness of the proposed approach. In future, we plan to adopt distributed optimization to speed up the learning process. We also plan to further extend the subspace partial view learning to nonlinear cases.

\begin{acks}
This work is supported by the National Natural Science Foundation of China (No. 62176270).
\end{acks}

\bibliographystyle{abbrv}
\bibliography{main}

\end{document}